\pdfoutput=1

\documentclass[11pt]{article}

\usepackage[final]{coling}

\usepackage{times}
\usepackage{latexsym}

\usepackage[T1]{fontenc}

\usepackage[utf8]{inputenc}

\usepackage{microtype}

\usepackage{inconsolata}

\usepackage{graphicx}

\usepackage{subcaption}
\usepackage{xcolor}
\usepackage{float} 
\usepackage{tabularx} 
\usepackage[most]{tcolorbox}

\usepackage{amsmath}
\usepackage{multirow}
\usepackage{graphicx}

\title{The First Multilingual Model For The Detection of Suicide Texts}

\author{Rodolfo Zevallos \\ Barcelona Supercomputing Center \\ rodolfo.zevallos@bsc.es
        \AND
         Annika Schoene \\ Northeastern University \\ a.schoene@northeastern.edu \And
         John E. Ortega \\ Northeastern University  \\ j.ortega@northeastern.edu}

\begin{document}
\maketitle
\begin{abstract}
Suicidal ideation is a serious health problem affecting millions of people worldwide. Social networks provide information about these mental health problems through users' emotional expressions. We propose a multilingual model leveraging transformer architectures like mBERT, XML-R, and mT5 to detect suicidal text across posts in six languages - \textit{Spanish}, \textit{English}, \textit{German}, \textit{Catalan}, \textit{Portuguese} and \textit{Italian}. A Spanish suicide ideation tweet dataset was translated into five other languages using SeamlessM4T. Each model was fine-tuned on this multilingual data and evaluated across classification metrics. Results showed mT5 achieving the best performance overall with F1 scores above 85\%, highlighting capabilities for cross-lingual transfer learning. The English and Spanish translations also displayed high quality based on perplexity. Our exploration underscores the importance of considering linguistic diversity in developing automated multilingual tools to identify suicidal risk. Limitations exist around semantic fidelity in translations and ethical implications which provide guidance for future human-in-the-loop evaluations.
\end{abstract}

\section{Introduction}

According to data published by the World Health Organization (WHO), over 700,000 people die by suicide each year \cite{world2021suicide}, with an additional 10 to 20 million attempting to take their own lives. Suicidal behavior typically begins with thoughts and ideations of death, eventually leading to suicide attempts - conscious acts with the purpose of ending one's existence \cite{liu2014life}. In this context, social networks have become spaces where individuals often disclose emotions and information that they don't feel comfortable sharing with healthcare providers \cite{ji2020suicidal,desmet2013emotion,sueki2015association}.

Early identification of signs of suicidal ideation in these online environments poses a major challenge. This is where Natural Language Processing (NLP) and Deep Learning (DL) can play a crucial role in the automatic detection of suicidal thoughts in computational settings. Furthermore, these computational approaches may contribute to the development of tools for harm reduction and prevention. However, the majority of research on suicidal ideation detection has been conducted on \textit{English} language data, resulting in a scarcity of linguistic resources (e.g.: datasets, lexicons, and Language Models) for most other languages.

Previous computational approaches to identifying suicidal ideation have relied heavily on hand-engineered features and domain expertise. For example, some studies have used structural and emotional features to train statistical prediction models on suicide text \cite{jones2007development,pestian2012s}. Additionally, conventional machine learning algorithms like Logistic Regression (LR) \cite{ramirez2020detection,jain2019machine,schoene2016automatic,o2015detecting}, Decision Tree (DT) \cite{jain2019machine,huang2015topic}, Naive Bayes (NB) \cite{shah2020hybridized,rabani2020detection,chiroma2018suiciderelated,schoene2016automatic}, Support Vector Machine (SVM) \cite{renjith2022ensemble,shah2020hybridized,ramirez2020detection}, K-nearest neighbor algorithm (KNN) \cite{shah2020hybridized,vioules2018detection} and Extreme Gradient Boost (XGBoost) \cite{rajesh2020suicidal,jain2019machine,ji2018supervised} have been applied. While achieving some success, these methods depend on costly feature engineering and professional knowledge.

Recently, deep learning has emerged as a promising approach that can automatically learn representations from data \cite{goldberg2022neural}. Moreover, deep learning techniques like CNNs \cite{yao2020detection,renjith2022ensemble,tadesse2019detection}, LSTMs \cite{haque2022comparative,tadesse2019detection,renjith2022ensemble,ji2018supervised,ma2018targeted}, BiLSTM \cite{haque2022comparative,zhang2022automatic,he2016pairwise} and DLSTMAttention \cite{zhang2022automatic,renjith2022ensemble} have been applied in detecting suicidal ideation, with competitive performance.

On the other hand, with the increasing use of pre-trained language models such as BERT \cite{devlin-etal-2019-bert}, RoBERTa \cite{liu2019roberta}, and mT5 \cite{xue2020mt5}, the landscape of suicide ideation detection has evolved significantly. These pre-trained models, developed through massive unsupervised learning on diverse linguistic tasks, offer a powerful foundation for understanding intricate nuances of language. Researchers are now exploring the adaptation of those models for the detection of suicidal ideation \cite{bhaumik2023mindwatch,devika2023bert,haque2020transformer}. Leveraging the contextual understanding encoded in these pre-trained models, studies have reported promising results in discerning subtle and complex expressions related to suicidal thoughts, contributing to the advancement of automated detection systems \cite{bhaumik2023mindwatch}. This shift towards pre-trained language models signifies a paradigmatic enhancement in the field, as it allows for a more nuanced comprehension of linguistic patterns associated with suicidal ideation, thereby enhancing the overall accuracy and sensitivity of detection algorithms.

In our approach, we emphasize the importance of addressing the detection of suicidal texts in the context of using multilingual language models. Translating a corpus from \textit{Spanish} into five different languages and fine-tuning a multilingual language model allows us to classify suicidal texts in various languages, thus expanding the applicability of our approach.

In our research, we aim to explore the effectiveness of multilingual pretrained language models such as mBERT \cite{devlin-etal-2019-bert}, XML-R \cite{liu2019roberta} and mT5 \cite{xue2020mt5} for detecting suicidal text on social media. The main focus of our study is to leverage these multilingual pretrained models to translate and recognize suicidal text in six languages: \textit{Spanish}, \textit{English}, \textit{German}, \textit{Catalan}, \textit{Portuguese} and \textit{Italian}.

Our primary contribution lies in the implementation of a multilingual language model with the capability to detect suicidal text in these six distinct languages. Additionally, to address the lack of labeled suicidal text in other languages, we utilize a labeled corpus and translate it into five different languages using an automatic translation model (SeamlessM4T\cite{barrault2023seamless}). Therefore, this approach enables us to effectively tackle the linguistic diversity present on social media and provides a valuable tool for the early identification of suicidal content in various cultural and linguistic contexts.

The main objectives of our study are:

\begin{enumerate}
    \item \textbf{Prediction of Suicidal Text in Six Languages:}
    The model focuses on predicting posts with suicidal content by analyzing the words or phrases written by users, utilizing multilingual pretrained language models such as mBERT, XML-R and mT5.
    
    \item \textbf{Improvement of Prediction Accuracy in Various Languages:}
    We aim to enhance the accuracy of predicting suicidal text by incorporating attention mechanisms from multilingual pretrained language models. These attention mechanisms highlight crucial aspects within the obtained information, providing effective detection in six different languages.
\end{enumerate}

The significant contributions of our work include:

\begin{enumerate}
    \item \textbf{Detection of Suicidal Texts Using Multilingual Pretrained Language Models (mBERT, XML-R, mT5):}
    We propose a model that integrates multilingual pretrained language models, including mBERT, XML-R amd mT5, for effective detection of suicidal texts in social media posts in six different languages.
    
    \item \textbf{Prediction of User-Specific Suicidal Tendencies in Various Languages:}
    The model examines the posts of specific users to determine if they exhibit suicidal tendencies, leveraging the capabilities of the mentioned multilingual pretrained language models.
\end{enumerate}

\section{Related Work}
The initial approaches to automatic suicide risk detection were based on identifying specific language features present in psychiatric literature. For instance, in \citet{lumontod2020seeing,tadesse2019detection}, the LIWC dictionary was used to extract emotional and cognitive markers, while \citet{masuda2013suicide} designed a set of emotional features such as feelings of loneliness, helplessness, and hopelessness. Additionally, \citet{pestian2010suicide} employed suicide notes to identify common language themes and styles.

However, the limitations of manual feature engineering in terms of scalability and adaptability have led to the exploration of more recent approaches based on deep learning, specifically pre-trained language models. In a comparative study \cite{tavchioski2023detection,sawhney2018exploring}, BERT, RoBERTa, BERTweet, and mentalBERT were evaluated on a Reddit dataset, revealing that pre-trained models consistently outperformed traditional classifiers \cite{Valeriano2020,maalouf2011logistic,aladaug2018detecting}.

In summary, pre-trained language models have shown promising results, often outperforming traditional methods in automatic suicide risk detection. However, most studies have been limited to relatively small datasets. Regarding linguistic diversity, studies have predominantly focused on English data from platforms such as Twitter \cite{kabir2023deptweet,coppersmith2015clpsych}, Reddit \cite{tavchioski2023detection,losada2016test,losada2017clef}, and Facebook. Nevertheless, no research has been found exploring multilingual language models for suicide risk detection. Models like mBERT \cite{devlin-etal-2019-bert}, XLM-RoBERTa \cite{liu2019roberta}, and mT5 \cite{xue2020mt5}, trained on multilingual data, could transfer linguistic knowledge across related languages, improving performance in low-resource situations for languages with less training data.

As far as is known, there are also no studies utilizing automatically translated datasets to leverage data from other languages. The quality of automatic translations of datasets from a source language to a target language could be crucial in increasing dataset size and improving the performance of trained models.

Both research directions, i.e., multilingual models and automatic translation of data, represent promising yet unexplored areas for automatic suicide risk detection, opening opportunities for significant contributions in this field.

\section{Experiments}
In this section, we delineate the setup of diverse experiments aimed at exploring the feasibility of a multilingual model capable of classifying suicidal texts across six different languages.

\subsection{Dataset}

The dataset we utilized in our experiments is the set of 2,068 Spanish tweets introduced in \citet{Valeriano2020}. This dataset was compiled by the authors through targeted keyword searches on expressions of suicidal ideation. The tweets were then manually annotated by humans, labeling each as either containing suicidal intent or not – a binary classification scheme. After annotation, the dataset contains 498 tweets (24\%) expressing suicidal ideas, with example phrases like "I want to disappear" or "I can't stand life anymore." The remaining 1,570 Spanish tweets do not express suicide risk.

We split the full dataset into training, validation. 80\% of the data, encompassing 1,654 Spanish tweets, was used for model training to learn signals of suicidal intent. The validation set makes up 20\% of the data, with 414 tweets, which was leveraged during model development for hyperparameter tuning and performance checks. Moreover, we used as test set the Lexicography Saves Lives (LSL) \citet{schoene2025}.

We leveraged this dataset by machine translating the entire corpus of 2,068 Spanish tweets into five other languages: Catalan, English, German, Italian, and Portuguese. The translations were produced using Facebook's SeamlessM4T model \cite{barrault2023seamless}, allowing us to obtain versions of the suicide texts dataset across multiple languages stemming from the original Spanish source data \cite{Valeriano2020}.

\subsection{Pre-trained language models}

The recent advances in neural network-based language models have demonstrated substantial improvements across a wide range of natural language processing tasks \cite{goldberg2022neural}. In particular, the introduction of Transformer architectures \cite{vaswani2017attention}   led to unprecedented progress in semantic and syntactic modeling capabilities. Unlike previous recurrent models such as LSTMs \cite{hochreiter1997long}, Transformer networks apply a purely attention-based mechanism to learn intricate context representations. By utilizing multiple attention heads in parallel, these architectures can capture both local and global dependencies in a sequence of tokens.

The original authors of the Transformer introduced a specific implementation called BERT \cite{devlin-etal-2019-bert}, which laid the groundwork for a new generation of contextualized language models. Through pre-training objectives such as predicting subsequent sentences and token masking, BERT achieves a deep syntactic and semantic understanding of language. However, the initial version of BERT was limited to the English language. Subsequent research focused on extending these models to a multilingual context to enable cross-lingual learning.

Adaptations such as mBERT\footnote{\url{https://github.com/google-research/bert/blob/master/multilingual.md}} \cite{devlin-etal-2019-bert} emerged, incorporating shared vocabularies and subword segmentation to represent a wide range of languages. Then, XML-R\footnote{\url{https://huggingface.co/xlm-roberta-base}} \cite{liu2019roberta} enhanced the multifaceted approach by adding byte-level tokenization and techniques like Whole-Word Masking. Finally, mT5\footnote{\url{https://github.com/google-research/multilingual-t5}}  \cite{xue2020mt5} adopted an encoder-decoder architecture instead of the exclusively encoder format. Considering the rapid progress in multilingual language models, this work aimed to evaluate three transformative alternatives for the automatic detection of suicidal ideation: mBERT, XML-R, and mT5. Through thorough experimentation, the goal is to determine their capabilities in both language and semantics.

Each chosen model presents unique characteristics, as described earlier, that could positively impact their performance for the given task. Additionally, all of them were pretrained in various languages, incorporating millions of trainable parameters and state-of-the-art techniques to enhance cross-linguistic transfer. In combination, this diversity allows addressing the problem from multiple perspectives, enabling a comprehensive evaluation of the relative advantages of different cutting-edge approaches for such a sensitive scenario as the expression of suicidal intentions.

For this study, three pre-trained language models were utilized and we oultine below further details about the architecture, hyperparameters, and training datasets for each.







\subsection{Suicide phrase recognition}

In the pursuit of robust multilingual performance, our experiments enlisted the capabilities of four cutting-edge language models: mBERT \cite{devlin-etal-2019-bert}, XML-R \cite{liu2019roberta} and mT5 \cite{xue2020mt5}. To fortify their adaptability, each model underwent a meticulous fine-tuning process. Leveraging the Spanish dataset, as previously detailed, and its translations into six languages—Catalan, English, German, Italian, and Portuguese—we aimed to comprehensively capture the nuances of suicidal text across linguistic variations.

The initial configurations for fine-tuning were aligned with the recommended settings provided by each language model. Subsequently, recognizing the intricate interplay of hyperparameters in influencing model performance, we conducted an exhaustive search to identify the most effective and contextually relevant hyperparameter sets for each individual model. This process was undertaken with a dual purpose: ensuring optimal performance across languages and tailoring the models to the specific intricacies of suicidal text classification.

We fine-tune mBERT, XML-R and mT5 on 1 NVIDIA 4070 GPUs with FP32. Model hyper-parameters are tuned on the validation set, where learning rate \{2e-5, 3e-5, 3e-5\}, batch size \{16, 16, 32\}, a dropout rate of \{0.3, 0.5, 0.5\}, a weight decay of 0.01, a warmup proportion of 0.01. For clarity and replicability, the detailed configurations for all models, including the identified hyperparameter sets, are meticulously documented in Table \ref{hyp}.

\begin{table}[]
\begin{tabular}{l|ccc}
Parameter                                                          & \multicolumn{1}{l}{\textbf{mBERT}} & \multicolumn{1}{l}{\textbf{XML-R}} & \multicolumn{1}{l}{\textbf{mT5}} \\ \hline
\begin{tabular}[c]{@{}l@{}}Starting\\ learning\\ rate\end{tabular} & 2e-5                               & 3e-5                                     & 3e-5                             \\
Batch size                                                         & 16                                 & 16                                       & 32                               \\
Epochs                                                             & 10                                 & 10                                       & 10                               \\
Dropout                                                            & 0.3                                & 0.5                                      & 0.5                              \\
\begin{tabular}[c]{@{}l@{}}Weight\\ decay\end{tabular}             & 0.01                               & 0.01                                     & 0.01                             \\
Optimizer                                                          & AdamW                              & AdamW                                    & AdamW                            \\ \hline
\end{tabular}
\caption{Hyperparameters for models fine-tuning}
\label{hyp}
\end{table}

\begin{table*}[]
\centering
\begin{tabular}{l|ccc|ccc|ccc}
\hline
\multirow{2}{*}{\textbf{Lang.}} & \multicolumn{3}{c|}{\textbf{mBERT}} & \multicolumn{3}{c|}{\textbf{XML-R}} & \multicolumn{3}{c}{\textbf{mT5}} \\ \cline{2-10} 
    & Acc.     & F1.   & AUC  & Acc.       & F1.     & AUC   & Acc.     & F1.   & AUC  \\ \hline
                                
Spanish                         & 82.4   & 82.1 & 82.2 & 84.6  & 84.3   & 84.3  & 87.9   & 87.7 & 87.8 \\
English                         & 83.6   & 83.3 & 83.1 & 85.6     & 85.5   & 85.4  & 88.5 &  88.1 & 88.1 \\
Italian                         & 78.7   & 78.5 & 78.4 & 80.6     & 80.6   & 80.4  & 83.3 &  83.2 & 83.1 \\
German                          & 81.1   & 80.9 & 80.7 & 82.9      & 82.9   & 82.8  & 86.2  & 86.1 & 86.0 \\
Catalan                         & 81.3   & 81.1 & 81.0 & 82.8    & 82.7   & 82.6  & 86.2 &  86.1 & 86.0 \\
Portuguese                      & 79.9   & 79.9 & 79.8 & 81.7   & 81.7   & 81.5  & 84.9 &  84.8 & 84.8 \\ \hline
\end{tabular}
\caption{Experimental results with mBERT, XML-R, and mT5 across different languages. Notation: Acc. = accuracy and F1. = F1-score.}
\label{tab:f1-scores}
\end{table*}

\section{Results}

In this section, we present an analysis of the results obtained from our fine-tuned language models—mBERT, XML-R and mT5 deployed in the task of suicidal text classification across six languages. Our objective is to scrutinize the models' performance intricacies, assess their multilingual adaptability, and glean insights into the efficacy of our approach.

\subsection{Classifiers Performance Analysis}

We delve into the nuanced evaluation of our language models' performance across six languages: Spanish, Catalan, English, German, Italian, and Portuguese. Table \ref{tab:f1-scores} shows that, mT5 displays superior performance over the other two models across all metrics and for all languages. The precision, recall, F1, and AUC scores are consistently high, surpassing 85\% in most cases.

This indicates that mT5 is exceptionally good at both positively detecting relevant cases (high recall) as well as minimizing false positives (high precision). It also maintains an adequate balance between both goals, as shown by its high F1-score. There is clearly substantial superiority of mT5 at this task compared to more generic BERT models.

On the other hand, we see that mBERT obtains the lowest scores, although still decent (around 80-83\% for key metrics). XML-R improves upon mBERT's results across all languages, suggesting that language-specific pretraining can be beneficial.

Regarding languages, English and Spanish consistently achieve the top scores across all models, followed by German and Catalan. Italian and Portuguese appear to be the most difficult. This could be due to several factors: data availability, similarity to English, etc.

An interesting finding is that the relative gaps between models remain remarkably stable across languages. This implies that the inherent strengths of each model transcend linguistic particularities. While some languages are more complex, all benefit from mT5's architectural improvements over BERT models.

In summary, mT5 is better suited to suicide text detection, especially excelling for English and Spanish. mBERT may perform adequately as a baseline, but there is clear room for improvement with more advanced models such as XML-R and especially mT5.

\begin{figure*}[t]
\centering
\includegraphics[width=0.8\textwidth]{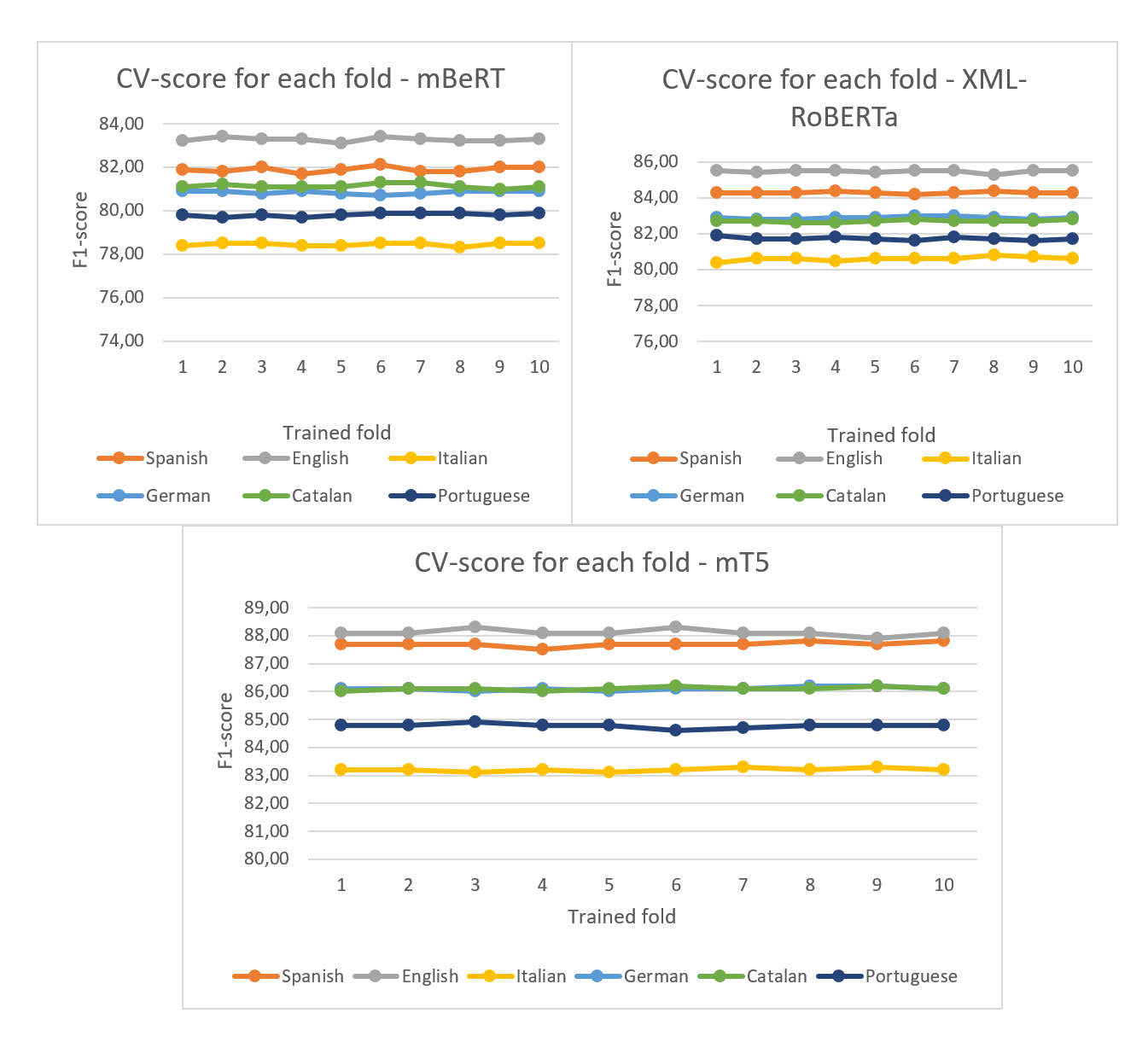} 
\caption{10-fold Cross-Validation for each language model}
\label{fig2}
\end{figure*}

\subsection{Model Validation}
To delve deeper into understanding the learning mechanism, we implemented k-fold cross-validation to determine the mean accuracy in our three models: mBERT, XML-R and mT5. Cross-validation is a widely used data resampling strategy to assess the generalization capabilities of predictive models and estimate the true estimation error.
In k-fold cross-validation, the learning set is divided into k subgroups of approximately equal length, and the number of subgroups produced is referred to as `fold'. This partition is achieved by randomly selecting examples from the learning set without replacement. Our language models, including mBERT, XML-R and mT5, were fine-tuned using k = 10 subsets representing the entire training set. Each model was then applied to the remaining subset, known as the validation set, and its performance was evaluated. This process was repeated until all k subsets had served as validation sets.

Subsequently, we proceeded to conduct additional tests in our six languages since our models are multilingual. This variant involves applying our models in scenarios with various languages, adding an additional level of complexity and versatility to the evaluation of their performance in detecting suicidal text. Figure \ref{fig2} illustrates the F1-score of each of our three language models for each fold in the cross-validation, highlighting their adaptability to diverse subsets of data and linguistic scenarios. This meticulous approach ensures robust training and optimal performance for each of our models in the detection of suicidal text, considering both linguistic diversity and the specific characteristics of mBERT, XML-R and mT5 in this particular context.

\section{Translation analysis}

For the translation of the Spanish dataset into the other 5 target languages (English, Catalan, German, Italian and Portuguese), this study employed the SeamlessM4T\footnote{\url{https://github.com/facebookresearch/seamless_communication}}   model developed by Facebook \cite{barrault2023seamless}.

SeamlessM4T is based on the Transformer architecture, demonstrating the effectiveness of cross-lingual model pretraining and transfer learning. In particular, it leverages a sequence-to-sequence model with encoder-decoder structure trained on large-scale data across multiple languages (100 languages).

The key advantages of this specific architecture include:

\begin{itemize}
    \item Attention-based interactions model both global and local dependencies in input and output sequences. This provides greater context and reduces reliance on recurrence.
    \item Multi-head self-attention combines representations from different positional offsets, learning synergistic features.
    \item Masked language modeling and denoising objectives during pretraining further enhance context modeling.
\end{itemize}

It was pretrained on a variety of language pairs, including Spanish, English, Catalan, Italian and Portuguese. It demonstrated excellent BLEU metrics on translations between these languages, corroborating its suitability for the present cross-lingual research task.

\subsection{Evaluation Metrics}

Perplexity serves as an indicator to quantify the quality of each translation. We employed monolingual language models specific to each target language, assessing their ability to predict word sequences in the translated texts.

\subsection{Results}

The perplexity scores for each translation are presented in the Table \ref{tab:translation-perplexity}: 

\begin{table}[h]
  \centering
  \begin{tabular}{l|c}
    \textbf{Translation to Language} & \textbf{Perplexity Score} \\
    \hline
    English\footnotemark[1] & 3.43 \\
    Catalan\footnotemark[2] & 5.65 \\
    German\footnotemark[2] & 8.18 \\
    Italian\footnotemark[2] & 7.75 \\
    Portuguese\footnotemark[2] & 4.61 \\
  \end{tabular}
  \caption{Perplexity scores for each translation.}
  \label{tab:translation-perplexity}
\end{table}

\footnotetext[1]{\url{https://huggingface.co/roberta-large}}
\footnotetext[2]{\url{https://huggingface.co/facebook/xlm-roberta-xl}}

With a low perplexity of 3.43, the English translation demonstrates notable coherence and fluency, suggesting successful adaptation from the Spanish source. This indicates that the model competently encoded the linguistic intricacies in mapping between the closely-related languages.

The Catalan translation perplexity of 5.65 signifies adequate synchronization from Spanish, though with slightly heightened linguistic complexity. This points to competent cross-lingual transfer learning while highlighting some incremental challenges for the more distant language pair.

However, the German translation perplexity of 8.18 underlines particular difficulties in adaptation, as substantiated by amplified linguistic complexity. As Spanish and German topologically diverge, this outcome spotlights the obstacles for conversion across more disparate languages.

Italian translation returned a perplexity of 7.75, underscoring reasonably effective adaptation despite lower fluency compared to other counterparts. This demonstrates capable inter-lingual transfer learning for the language pair, albeit with some decline in conversion quality.

Finally, the Portuguese translation perplexity of 4.61 reflects adept transformation from Spanish, mirroring the performance benchmark set by English. The proximity between Spanish and Portuguese facilitates smooth cross-lingual mapping, resulting in harmonized coherence.

\section{Discussion}

This study explores the use of pre-trained multilingual language models, including mBERT, XML-R, and mT5, for the automatic detection of suicidal texts in social media posts across six languages: Spanish, English, German, Catalan, Portuguese, and Italian. The results show mT5 achieving the best performance overall, with F1 scores above 85\%, highlighting capabilities for cross-lingual transfer learning.

An interesting finding is that the relative gaps between models remain remarkably stable across languages. This implies that the inherent strengths of each model transcend linguistic particularities. While some languages are more complex, all benefit from mT5’s architectural improvements over BERT models.

Regarding limitations, direct extrapolation of the results to other languages must be approached cautiously, given the wide linguistic diversity and potential impact of cultural nuances on interpreting suicidal texts. Furthermore, the quality of translations and, consequently, the predictive model, is inherently tied to the effectiveness of pre-trained models, indicating a constant need for improvements in this area.

While this study presents a promising model for multilingual detection of suicidal texts, there are several directions to extend and strengthen this line of research. Some of these include: expanding linguistic scope by incorporating a broader spectrum of languages; enriching training data with more instances and diversity of sources; using specialized metrics to quantify the usefulness of the early detection model; and implementation of a user-friendly interface enabling integration into healthcare settings.

In summary, the focus on multilingual translation emerges as a crucial step in constructing an effective predictive model for suicidal texts across six languages. The identified conclusions and limitations provide guidance for future developments, emphasizing the need for linguistic and cultural considerations.

\section{Conclusion}
In the pursuit of a predictive model for suicidal texts in six languages, our exploration into multilingual translation yields critical insights. We observe that translations into \textit{English} and \textit{Portuguese} excel, showcasing the ability to preserve intent and coherence in sensitive contexts such as suicidal content.

Sensitivity to linguistic diversity emerges as a pivotal element in this process. While synchronization in translations into Catalan was acceptable, adaptations into German and Italian posed challenges, underscoring the importance of considering linguistic nuances in constructing a robust predictive model. The versatility of multilingual models, especially mT5, proves to be a valuable resource in this scenario. These models demonstrate a remarkable ability to maintain the integrity of suicidal content across diverse languages, providing a solid foundation for building a multilingual predictive model. Automated evaluation, though guided by objective metrics such as perplexity, does not replace human assessment for sensitivity and semantic fidelity in suicidal content. The implementation of human evaluations in subsequent phases is essential to ensure the appropriateness and ethical considerations of the model.

In summary, our focus on multilingual translation emerges as a crucial step in constructing a predictive model for suicidal texts in six languages. The identified conclusions and limitations provide guidance for future developments, emphasizing the need for linguistic and cultural considerations, as well as continuous improvements in pre-trained models and human evaluations to achieve an effective and ethical model.

\section{Ethical Considerations}

There are a number of aspects to consider when using pretrained language models to automatically translate suicide related language, especially given the sensitive nature of the data. Firstly, we have to consider user privacy and be aware of the impact online surveillance, collection of sensitive data and people's health. Furthermore, there are concerns around linguistic, cultural and contextual accuracy when automatically translating suicide-related tweets, where there can be issues around accurate translations and misrepresentation of cultural or conceptual concepts. Finally,

\section{Limitations and Future Work}

Direct extrapolation of our results to other languages must be approached cautiously, given the wide linguistic diversity and the potential impact of cultural nuances on the interpretation of suicidal texts. Furthermore, the quality of translations and, consequently, the predictive model, is inherently linked to the effectiveness of pre-trained models, indicating a constant need for improvements in this area.

While this study presents a promising model for multilingual detection of suicidal ideation, there are several directions to extend and strengthen this line of research:

\begin{itemize}

    \item \textbf{Expansion of Linguistic Scope} Incorporating a broader spectrum of languages would be key to achieving a globally impactful tool. Languages with limited use of digital technologies like \textit{Hindi}, \textit{Arabic} or \textit{Chinese} pose challenges due to scarce representation in training data. Techniques such as small-scale automatic translation of annotated data and adaptation of models to new languages through transfer learning could help bridge this gap.

    \item \textbf{Enrichment of Training Data} Having more instances and diversity of sources in the initial Spanish dataset would enhance derived models. Collecting content from platforms like Reddit \cite{zirikly-etal-2019-clpsych} and Facebook \cite{ophir2020deep} with a higher prevalence of mental health themes could be beneficial. Expanding labels to capture emotional nuances, linguistic subtleties and a more granular view of suicide-realted content (e.g.: moving beyond binary classification) could also contribute.

    \item \textbf{Specialized Metrics} To more precisely quantify the utility of the early detection model, metrics like average latency to high-risk posts or rate of early false negatives should be incorporated. Establishing how these indicators vary across dialectal and sociocultural differences is essential.


    \item \textbf{Implementation for Healthcare Institutions} Developing a user-friendly interface for  models that enables integration in healthcare settings would ease the transition of this technology into real-world applications. Achieving integration with existing clinical record systems and care workflows could further its adoption.

\end{itemize}

Addressing these extensions would provide a comprehensive system with superior accuracy, broad multilingual reach and significant impact on the timely detection and prevention of suicidal behaviors through computing.

\bibliography{custom}

\begin{thebibliography}{50}
\providecommand{\natexlab}[1]{#1}

\bibitem[{Alada{\u{g}} et~al.(2018)Alada{\u{g}}, Muderrisoglu, Akbas, Zahmacioglu, and Bingol}]{aladaug2018detecting}
Ahmet~Emre Alada{\u{g}}, Serra Muderrisoglu, Naz~Berfu Akbas, Oguzhan Zahmacioglu, and Haluk~O Bingol. 2018.
\newblock Detecting suicidal ideation on forums: proof-of-concept study.
\newblock \emph{Journal of medical Internet research}, 20(6):e9840.

\bibitem[{Barrault et~al.(2023)Barrault, Chung, Meglioli, Dale, Dong, Duppenthaler, Duquenne, Ellis, Elsahar, Haaheim et~al.}]{barrault2023seamless}
Lo{\"\i}c Barrault, Yu-An Chung, Mariano~Coria Meglioli, David Dale, Ning Dong, Mark Duppenthaler, Paul-Ambroise Duquenne, Brian Ellis, Hady Elsahar, Justin Haaheim, et~al. 2023.
\newblock Seamless: Multilingual expressive and streaming speech translation.
\newblock \emph{arXiv preprint arXiv:2312.05187}.

\bibitem[{Bhaumik et~al.(2023)Bhaumik, Srivastava, Jalali, Ghosh, and Chandrasekaran}]{bhaumik2023mindwatch}
Runa Bhaumik, Vineet Srivastava, Arash Jalali, Shanta Ghosh, and Ranganathan Chandrasekaran. 2023.
\newblock Mindwatch: A smart cloud-based ai solution for suicide ideation detection leveraging large language models.
\newblock \emph{medRxiv}, pages 2023--09.

\bibitem[{Chiroma et~al.(2018)Chiroma, Liu, and Cocea}]{chiroma2018suiciderelated}
Fatima Chiroma, Han Liu, and Mihaela Cocea. 2018.
\newblock Suiciderelated text classification with prism algorithm.
\newblock In \emph{2018 International Conference on Machine Learning and Cybernetics (ICMLC)}, volume~2, pages 575--580. IEEE.

\bibitem[{Coppersmith et~al.(2015)Coppersmith, Dredze, Harman, Hollingshead, and Mitchell}]{coppersmith2015clpsych}
Glen Coppersmith, Mark Dredze, Craig Harman, Kristy Hollingshead, and Margaret Mitchell. 2015.
\newblock Clpsych 2015 shared task: Depression and ptsd on twitter.
\newblock In \emph{Proceedings of the 2nd workshop on computational linguistics and clinical psychology: from linguistic signal to clinical reality}, pages 31--39.

\bibitem[{Desmet and Hoste(2013)}]{desmet2013emotion}
Bart Desmet and V{\'e}Ronique Hoste. 2013.
\newblock Emotion detection in suicide notes.
\newblock \emph{Expert Systems with Applications}, 40(16):6351--6358.

\bibitem[{Devika et~al.(2023)Devika, Pooja, Arpitha, and Vinayakumar}]{devika2023bert}
SP~Devika, MR~Pooja, MS~Arpitha, and Ravi Vinayakumar. 2023.
\newblock Bert-based approach for suicide and depression identification.
\newblock In \emph{Proceedings of Third International Conference on Advances in Computer Engineering and Communication Systems: ICACECS 2022}, pages 435--444. Springer.

\bibitem[{Devlin et~al.(2019)Devlin, Chang, Lee, and Toutanova}]{devlin-etal-2019-bert}
Jacob Devlin, Ming-Wei Chang, Kenton Lee, and Kristina Toutanova. 2019.
\newblock \href {https://doi.org/10.18653/v1/N19-1423} {{BERT}: Pre-training of deep bidirectional transformers for language understanding}.
\newblock In \emph{Proceedings of the 2019 Conference of the North {A}merican Chapter of the Association for Computational Linguistics: Human Language Technologies, Volume 1 (Long and Short Papers)}, pages 4171--4186, Minneapolis, Minnesota. Association for Computational Linguistics.

\bibitem[{Goldberg(2022)}]{goldberg2022neural}
Yoav Goldberg. 2022.
\newblock \emph{Neural network methods for natural language processing}.
\newblock Springer Nature.

\bibitem[{Haque et~al.(2020)Haque, Nur, Al~Jahan, Mahmud, and Shah}]{haque2020transformer}
Farsheed Haque, Ragib~Un Nur, Shaeekh Al~Jahan, Zarar Mahmud, and Faisal~Muhammad Shah. 2020.
\newblock A transformer based approach to detect suicidal ideation using pre-trained language models.
\newblock In \emph{2020 23rd international conference on computer and information technology (ICCIT)}, pages 1--5. IEEE.

\bibitem[{Haque et~al.(2022)Haque, Islam, Islam, and Ahsan}]{haque2022comparative}
Rezaul Haque, Naimul Islam, Maidul Islam, and Md~Manjurul Ahsan. 2022.
\newblock A comparative analysis on suicidal ideation detection using nlp, machine, and deep learning.
\newblock \emph{Technologies}, 10(3):57.

\bibitem[{He and Lin(2016)}]{he2016pairwise}
Hua He and Jimmy Lin. 2016.
\newblock Pairwise word interaction modeling with deep neural networks for semantic similarity measurement.
\newblock In \emph{Proceedings of the 2016 conference of the north American chapter of the Association for Computational Linguistics: human language technologies}, pages 937--948.

\bibitem[{Hochreiter and Schmidhuber(1997)}]{hochreiter1997long}
Sepp Hochreiter and J{\"u}rgen Schmidhuber. 1997.
\newblock Long short-term memory.
\newblock \emph{Neural computation}, 9(8):1735--1780.

\bibitem[{Huang et~al.(2015)Huang, Li, Zhang, Liu, Chiu, and Zhu}]{huang2015topic}
Xiaolei Huang, Xin Li, Lei Zhang, Tianli Liu, David Chiu, and Tingshao Zhu. 2015.
\newblock Topic model for identifying suicidal ideation in chinese microblog.
\newblock In \emph{Proceedings of the 29th pacific asia conference on language, information and computation}, pages 553--562. Waseda University.

\bibitem[{Jain et~al.(2019)Jain, Narayan, Dewang, Bhartiya, Meena, and Kumar}]{jain2019machine}
Swati Jain, Suraj~Prakash Narayan, Rupesh~Kumar Dewang, Utkarsh Bhartiya, Nalini Meena, and Varun Kumar. 2019.
\newblock A machine learning based depression analysis and suicidal ideation detection system using questionnaires and twitter.
\newblock In \emph{2019 IEEE students conference on engineering and systems (SCES)}, pages 1--6. IEEE.

\bibitem[{Ji et~al.(2020)Ji, Pan, Li, Cambria, Long, and Huang}]{ji2020suicidal}
Shaoxiong Ji, Shirui Pan, Xue Li, Erik Cambria, Guodong Long, and Zi~Huang. 2020.
\newblock Suicidal ideation detection: A review of machine learning methods and applications.
\newblock \emph{IEEE Transactions on Computational Social Systems}, 8(1):214--226.

\bibitem[{Ji et~al.(2018)Ji, Yu, Fung, Pan, and Long}]{ji2018supervised}
Shaoxiong Ji, Celina~Ping Yu, Sai-fu Fung, Shirui Pan, and Guodong Long. 2018.
\newblock Supervised learning for suicidal ideation detection in online user content.
\newblock \emph{Complexity}, 2018.

\bibitem[{Jones and Bennell(2007)}]{jones2007development}
Natalie~J Jones and Craig Bennell. 2007.
\newblock The development and validation of statistical prediction rules for discriminating between genuine and simulated suicide notes.
\newblock \emph{Archives of Suicide Research}, 11(2):219--233.

\bibitem[{Kabir et~al.(2023)Kabir, Ahmed, Hasan, Laskar, Joarder, Mahmud, and Hasan}]{kabir2023deptweet}
Mohsinul Kabir, Tasnim Ahmed, Md~Bakhtiar Hasan, Md~Tahmid~Rahman Laskar, Tarun~Kumar Joarder, Hasan Mahmud, and Kamrul Hasan. 2023.
\newblock Deptweet: A typology for social media texts to detect depression severities.
\newblock \emph{Computers in Human Behavior}, 139:107503.

\bibitem[{Liu and Miller(2014)}]{liu2014life}
Richard~T Liu and Ivan Miller. 2014.
\newblock Life events and suicidal ideation and behavior: A systematic review.
\newblock \emph{Clinical psychology review}, 34(3):181--192.

\bibitem[{Liu et~al.(2019)Liu, Ott, Goyal, Du, Joshi, Chen, Levy, Lewis, Zettlemoyer, and Stoyanov}]{liu2019roberta}
Yinhan Liu, Myle Ott, Naman Goyal, Jingfei Du, Mandar Joshi, Danqi Chen, Omer Levy, Mike Lewis, Luke Zettlemoyer, and Veselin Stoyanov. 2019.
\newblock Roberta: A robustly optimized bert pretraining approach.
\newblock \emph{arXiv preprint arXiv:1907.11692}.

\bibitem[{Losada and Crestani(2016)}]{losada2016test}
David~E Losada and Fabio Crestani. 2016.
\newblock A test collection for research on depression and language use.
\newblock In \emph{International conference of the cross-language evaluation forum for European languages}, pages 28--39. Springer.

\bibitem[{Losada et~al.(2017)Losada, Crestani, and Parapar}]{losada2017clef}
David~E Losada, Fabio Crestani, and Javier Parapar. 2017.
\newblock Clef 2017 erisk overview: Early risk prediction on the internet: Experimental foundations.
\newblock \emph{CLEF (Working Notes)}, 850.

\bibitem[{Lumontod~III(2020)}]{lumontod2020seeing}
Robinson~Z Lumontod~III. 2020.
\newblock Seeing the invisible: Extracting signs of depression and suicidal ideation from college students’ writing using liwc a computerized text analysis.
\newblock \emph{Int. J. Res. Stud. Educ}, 9:31--44.

\bibitem[{Ma et~al.(2018)Ma, Peng, and Cambria}]{ma2018targeted}
Yukun Ma, Haiyun Peng, and Erik Cambria. 2018.
\newblock Targeted aspect-based sentiment analysis via embedding commonsense knowledge into an attentive lstm.
\newblock In \emph{Proceedings of the AAAI conference on artificial intelligence}, volume~32.

\bibitem[{Maalouf(2011)}]{maalouf2011logistic}
Maher Maalouf. 2011.
\newblock Logistic regression in data analysis: an overview.
\newblock \emph{International Journal of Data Analysis Techniques and Strategies}, 3(3):281--299.

\bibitem[{Masuda et~al.(2013)Masuda, Kurahashi, and Onari}]{masuda2013suicide}
Naoki Masuda, Issei Kurahashi, and Hiroko Onari. 2013.
\newblock Suicide ideation of individuals in online social networks.
\newblock \emph{PloS one}, 8(4):e62262.

\bibitem[{O'dea et~al.(2015)O'dea, Wan, Batterham, Calear, Paris, and Christensen}]{o2015detecting}
Bridianne O'dea, Stephen Wan, Philip~J Batterham, Alison~L Calear, Cecile Paris, and Helen Christensen. 2015.
\newblock Detecting suicidality on twitter.
\newblock \emph{Internet Interventions}, 2(2):183--188.

\bibitem[{Ophir et~al.(2020)Ophir, Tikochinski, Asterhan, Sisso, and Reichart}]{ophir2020deep}
Yaakov Ophir, Refael Tikochinski, Christa~SC Asterhan, Itay Sisso, and Roi Reichart. 2020.
\newblock Deep neural networks detect suicide risk from textual facebook posts.
\newblock \emph{Scientific reports}, 10(1):16685.

\bibitem[{Organization et~al.(2021)}]{world2021suicide}
World~Health Organization et~al. 2021.
\newblock Suicide worldwide in 2019: global health estimates.

\bibitem[{Pestian et~al.(2010)Pestian, Nasrallah, Matykiewicz, Bennett, and Leenaars}]{pestian2010suicide}
John Pestian, Henry Nasrallah, Pawel Matykiewicz, Aurora Bennett, and Antoon Leenaars. 2010.
\newblock Suicide note classification using natural language processing: A content analysis.
\newblock \emph{Biomedical informatics insights}, 3:BII--S4706.

\bibitem[{Pestian et~al.(2012)Pestian, Matykiewicz, and Linn-Gust}]{pestian2012s}
John~P Pestian, Pawel Matykiewicz, and Michelle Linn-Gust. 2012.
\newblock What's in a note: construction of a suicide note corpus.
\newblock \emph{Biomedical informatics insights}, 5:BII--S10213.

\bibitem[{Rabani et~al.(2020)Rabani, Khan, and Khanday}]{rabani2020detection}
Syed~Tanzeel Rabani, Qamar~Rayees Khan, and Akib Mohi Ud~Din Khanday. 2020.
\newblock Detection of suicidal ideation on twitter using machine learning \& ensemble approaches.
\newblock \emph{Baghdad science journal}, 17(4):1328--1328.

\bibitem[{Rajesh~Kumar et~al.(2020)Rajesh~Kumar, Rama~Rao, Nayak, and Chandra}]{rajesh2020suicidal}
E~Rajesh~Kumar, KVSN Rama~Rao, Soumya~Ranjan Nayak, and Ramesh Chandra. 2020.
\newblock Suicidal ideation prediction in twitter data using machine learning techniques.
\newblock \emph{Journal of Interdisciplinary Mathematics}, 23(1):117--125.

\bibitem[{Ram{\'\i}rez-Cifuentes et~al.(2020)Ram{\'\i}rez-Cifuentes, Freire, Baeza-Yates, Punt{\'\i}, Medina-Bravo, Velazquez, Gonfaus, and Gonz{\`a}lez}]{ramirez2020detection}
Diana Ram{\'\i}rez-Cifuentes, Ana Freire, Ricardo Baeza-Yates, Joaquim Punt{\'\i}, Pilar Medina-Bravo, Diego~Alejandro Velazquez, Josep~Maria Gonfaus, and Jordi Gonz{\`a}lez. 2020.
\newblock Detection of suicidal ideation on social media: multimodal, relational, and behavioral analysis.
\newblock \emph{Journal of medical internet research}, 22(7):e17758.

\bibitem[{Renjith et~al.(2022)Renjith, Abraham, Jyothi, Chandran, and Thomson}]{renjith2022ensemble}
Shini Renjith, Annie Abraham, Surya~B Jyothi, Lekshmi Chandran, and Jincy Thomson. 2022.
\newblock An ensemble deep learning technique for detecting suicidal ideation from posts in social media platforms.
\newblock \emph{Journal of King Saud University-Computer and Information Sciences}, 34(10):9564--9575.

\bibitem[{Sawhney et~al.(2018)Sawhney, Manchanda, Mathur, Shah, and Singh}]{sawhney2018exploring}
Ramit Sawhney, Prachi Manchanda, Puneet Mathur, Rajiv Shah, and Raj Singh. 2018.
\newblock Exploring and learning suicidal ideation connotations on social media with deep learning.
\newblock In \emph{Proceedings of the 9th workshop on computational approaches to subjectivity, sentiment and social media analysis}, pages 167--175.

\bibitem[{Schoene et~al.(2025)Schoene, Ortega, Zevallos, and Ihle}]{schoene2025}
Annika Schoene, John~E. Ortega, Rodolfo~Joel Zevallos, and Laura Ihle. 2025.
\newblock Lexicography saves lives (lsl): Automatically translating suicide-related language.
\newblock In \emph{Proceedings of the 31st International Conference on Computational Linguistics}. International Committee on Computational Linguistics.

\bibitem[{Schoene and Dethlefs(2016)}]{schoene2016automatic}
Annika~Marie Schoene and Nina Dethlefs. 2016.
\newblock Automatic identification of suicide notes from linguistic and sentiment features.
\newblock In \emph{Proceedings of the 10th SIGHUM workshop on language technology for cultural heritage, social sciences, and humanities}, pages 128--133.

\bibitem[{Shah et~al.(2020)Shah, Haque, Nur, Al~Jahan, and Mamud}]{shah2020hybridized}
Faisal~Muhammad Shah, Farsheed Haque, Ragib~Un Nur, Shaeekh Al~Jahan, and Zarar Mamud. 2020.
\newblock A hybridized feature extraction approach to suicidal ideation detection from social media post.
\newblock In \emph{2020 IEEE Region 10 Symposium (TENSYMP)}, pages 985--988. IEEE.

\bibitem[{Sueki(2015)}]{sueki2015association}
Hajime Sueki. 2015.
\newblock The association of suicide-related twitter use with suicidal behaviour: a cross-sectional study of young internet users in japan.
\newblock \emph{Journal of affective disorders}, 170:155--160.

\bibitem[{Tadesse et~al.(2019)Tadesse, Lin, Xu, and Yang}]{tadesse2019detection}
Michael~M Tadesse, Hongfei Lin, Bo~Xu, and Liang Yang. 2019.
\newblock Detection of depression-related posts in reddit social media forum.
\newblock \emph{Ieee Access}, 7:44883--44893.

\bibitem[{Tavchioski et~al.(2023)Tavchioski, Robnik-{\v{S}}ikonja, and Pollak}]{tavchioski2023detection}
Ilija Tavchioski, Marko Robnik-{\v{S}}ikonja, and Senja Pollak. 2023.
\newblock Detection of depression on social networks using transformers and ensembles.
\newblock \emph{arXiv preprint arXiv:2305.05325}.

\bibitem[{Valeriano et~al.(2020)Valeriano, Condori-Larico, and Sulla-Torres}]{Valeriano2020}
Kid Valeriano, Alexia Condori-Larico, and Josè Sulla-Torres. 2020.
\newblock \href {https://doi.org/10.14569/IJACSA.2020.0110489} {Detection of suicidal intent in spanish language social networks using machine learning}.
\newblock \emph{International Journal of Advanced Computer Science and Applications}, 11(4).

\bibitem[{Vaswani et~al.(2017)Vaswani, Shazeer, Parmar, Uszkoreit, Jones, Gomez, Kaiser, and Polosukhin}]{vaswani2017attention}
Ashish Vaswani, Noam Shazeer, Niki Parmar, Jakob Uszkoreit, Llion Jones, Aidan~N Gomez, {\L}ukasz Kaiser, and Illia Polosukhin. 2017.
\newblock Attention is all you need.
\newblock \emph{Advances in neural information processing systems}, 30.

\bibitem[{Vioules et~al.(2018)Vioules, Moulahi, Az{\'e}, and Bringay}]{vioules2018detection}
M~Johnson Vioules, Bilel Moulahi, J{\'e}r{\^o}me Az{\'e}, and Sandra Bringay. 2018.
\newblock Detection of suicide-related posts in twitter data streams.
\newblock \emph{IBM Journal of Research and Development}, 62(1):7--1.

\bibitem[{Xue et~al.(2020)Xue, Constant, Roberts, Kale, Al-Rfou, Siddhant, Barua, and Raffel}]{xue2020mt5}
Linting Xue, Noah Constant, Adam Roberts, Mihir Kale, Rami Al-Rfou, Aditya Siddhant, Aditya Barua, and Colin Raffel. 2020.
\newblock mt5: A massively multilingual pre-trained text-to-text transformer.
\newblock \emph{arXiv preprint arXiv:2010.11934}.

\bibitem[{Yao et~al.(2020)Yao, Rashidian, Dong, Duanmu, Rosenthal, and Wang}]{yao2020detection}
Hannah Yao, Sina Rashidian, Xinyu Dong, Hongyi Duanmu, Richard~N Rosenthal, and Fusheng Wang. 2020.
\newblock Detection of suicidality among opioid users on reddit: machine learning--based approach.
\newblock \emph{Journal of medical internet research}, 22(11):e15293.

\bibitem[{Zhang et~al.(2022)Zhang, Schoene, and Ananiadou}]{zhang2022automatic}
Tianlin Zhang, Annika~M Schoene, and Sophia Ananiadou. 2022.
\newblock Automatic identification of suicide notes with a transformer-based deep learning model elsevier, vol. 25.

\bibitem[{Zirikly et~al.(2019)Zirikly, Resnik, Uzuner, and Hollingshead}]{zirikly-etal-2019-clpsych}
Ayah Zirikly, Philip Resnik, {\"O}zlem Uzuner, and Kristy Hollingshead. 2019.
\newblock \href {https://doi.org/10.18653/v1/W19-3003} {{CLP}sych 2019 shared task: Predicting the degree of suicide risk in {R}eddit posts}.
\newblock In \emph{Proceedings of the Sixth Workshop on Computational Linguistics and Clinical Psychology}, pages 24--33, Minneapolis, Minnesota. Association for Computational Linguistics.

\end{thebibliography}



\end{document}